\documentclass[12pt]{article}

\usepackage[utf8]{inputenc}
\usepackage[T1]{fontenc}
\usepackage{geometry}
\usepackage{setspace}
\usepackage{amsmath, amssymb, bm}
\usepackage{amsthm}

\theoremstyle{plain}
\newtheorem{theorem}{Theorem}
\newtheorem{lemma}[theorem]{Lemma}
\newtheorem{proposition}[theorem]{Proposition}
\newtheorem{corollary}[theorem]{Corollary}

\theoremstyle{definition}

\theoremstyle{remark}

\usepackage{algorithm}
\usepackage{algpseudocode}
\usepackage{tikz}
\usepackage{mdframed}
\usepackage{array,tabularx}
\usepackage{subcaption}
\usepackage{booktabs}

\usepackage{siunitx}
\usepackage{natbib}
\usepackage{endnotes}
\usepackage{hyperref}
\usepackage{graphicx}

\title{Multi-Agent Reinforcement Learning for Intraday Operating Rooms Scheduling under Uncertainty}

\author{
Kailiang Liu \\
{\small Department of Mathematics, National University of Singapore} \\
{\small \texttt{liukl@u.nus.edu}} 
\and
Ying Chen \\
{\small Department of Mathematics \& Risk Management Institute \& Center for Quantitative Finance,}\\
{\small National University of Singapore} \\
{\small \texttt{matcheny@nus.edu.sg}}
\and
Ralf Bornd\"orfer \\
{\small Department of Mathematics and Computer Science, Freie Universit\"at Berlin} \\
{\small Department of Network Optimization, Zuse Institute Berlin} \\
{\small \texttt{borndoerfer@zib.de}}
\and
Thorsten Koch \\
{\small Faculty of Mathematics and Natural Sciences, Technische Universit\"at Berlin} \\
{\small Department of Applied Algorithmic Methods, Zuse Institute Berlin} \\
{\small \texttt{koch@zib.de}}
}

\date{}

\begin{document}
\maketitle

\begin{abstract}
Intraday surgical scheduling is a multi-objective decision problem under uncertainty—balancing elective throughput, urgent and emergency demand, delays, sequence-dependent setups, and overtime. We formulate the problem as a cooperative Markov game and propose a multi-agent reinforcement learning (MARL) framework in which each operating room (OR) is an agent trained with centralized training and decentralized execution. All agents share a policy trained via Proximal Policy Optimization (PPO), which maps rich system states to actions, while a within-epoch sequential assignment protocol constructs conflict-free joint schedules across ORs. A mixed-integer pre-schedule provides reference starting times for electives; we impose type-specific quadratic delay penalties relative to these references and a terminal overtime penalty, yielding a single reward that captures throughput, timeliness, and staff workload. In simulations reflecting a realistic hospital mix (six ORs, eight surgery types, random urgent and emergency arrivals), the learned policy outperforms six rule-based heuristics across seven metrics and three evaluation subsets, and, relative to an ex post MIP oracle, quantifies optimality gaps. Policy analytics reveal interpretable behavior—prioritizing emergencies, batching similar cases to reduce setups, and deferring lower-value electives. We also derive a suboptimality bound for the sequential decomposition under simplifying assumptions. We discuss limitations—including OR homogeneity and the omission of explicit staffing constraints—and outline extensions. Overall, the approach offers a practical, interpretable, and tunable data-driven complement to optimization for real-time OR scheduling.
\end{abstract}


\section{Introduction}\label{sec:intro}

Operating rooms (ORs) are the financial and clinical nexus of the modern hospital. Surgical services are frequently reported to account for a large share of hospital finances---often \emph{60--70\% of revenue} and roughly \emph{35--40\% of costs}---placing extraordinary pressure on managers to deploy OR time efficiently while protecting access and outcomes \citep{Rothstein2018}. At the same time, 
the cost of operating room (OR) time typically ranges from approximately \$20 per minute to over \$100 per minute, with direct and indirect components that rise quickly when schedules run long \citep{Childers2018}. Compounding these pressures, same-day surgical cancellations---a common form of intraday disruption---occur at material rates (meta-analytic prevalence \emph{about 18\%} globally), eroding throughput and patient experience \citep{Abate2020}. Together, these facts underscore the managerial stakes of day-of-surgery decision making.

Intraday (day-of) OR scheduling is inherently a \emph{multi-objective} decision problem under pervasive \emph{uncertainty}. Managers must balance \emph{patient (case) throughput} and \emph{timely access}---especially for urgent and emergency demand---against \emph{patient waiting times}, \emph{OR idle time}, and \emph{staff overtime}, while sustaining \emph{revenue} and ensuring effective use of surgical teams and equipment. Even with a sound pre-day elective plan, day-of realities routinely erode static schedules: procedure durations are stochastic; urgent cases arrive unpredictably; emergency batches periodically materialize; and sequence-dependent setup work (cleaning, instrumentation, room turnover) creates frictions that amplify small delays \citep{Denton2007,SILVA2020102066}. These dynamics generate a large, evolving state space and a stream of time-critical, sequential decisions that are difficult to solve optimally in real time.

\paragraph{Approach.}
We address intraday OR scheduling as a \emph{sequential} decision problem and develop a modern learning-based solution rooted in \emph{multi-agent reinforcement learning} (MARL). We model the system as a \emph{cooperative Markov game} in which each OR is an agent acting toward a shared hospital objective. The design follows \emph{centralized training and decentralized execution} (CTDE): during training, a central learner has access to the global state to induce cooperative behavior; at run time, each OR executes the learned policy locally and in real time. We instantiate CTDE with \emph{Proximal Policy Optimization} (PPO), a policy-gradient method known for training stability \citep{schulman2017proximalpolicyoptimizationalgorithms}.

To produce \emph{conflict-free} joint decisions without enumerating exponentially large joint action spaces, we employ a \emph{within-epoch sequential assignment} mechanism: at each decision epoch, available ORs take turns (in a fixed or randomized order) selecting actions; later-acting ORs observe earlier choices, eliminating collisions such as assigning the same job twice. The policy observes a rich state summarizing time; room occupancy/progress; type-specific queues; and elective reference start times derived from a pre-day plan.

Crucial operational elements are embedded directly in the \emph{reward design}. First, we generate an \emph{elective pre-schedule} via a mixed-integer program (MIP), which provides reference start times\footnote{For \emph{non-elective patients}, the arrival time itself serves as the reference, ensuring that their waiting time is explicitly penalized in the reward.}; the reinforcement signal then imposes \emph{type-specific quadratic delay penalties} relative to those references, discouraging excessive waits and reflecting that marginal disutility rises with delay (particularly for higher-urgency classes). Second, we include \emph{sequence-dependent setup times} to capture cleaning and preparation overhead between heterogeneous case types; this incentivizes batching of similar types when clinically appropriate and aligns with well-known turnover sensitivities in OR practice. Third, a \emph{terminal overtime penalty} captures labor and downstream costs of running past regular hours (with implications for staff well-being too); idle penalties can also be incorporated to temper unnecessary slack, though they are not included in the reward function in this study. This single scalar reward induces policies that implicitly trade off \emph{throughput}, \emph{timeliness}, and \emph{workload/costs} without continual retuning.

\paragraph{Empirical setting and evaluation.}
We evaluate the approach in a high-fidelity hospital simulation that reflects a realistic OR setting: six parallel rooms; eight surgery types with diverse duration distributions; and a stochastic mix of elective, urgent, and emergency arrivals over an intraday horizon of 100 discrete time units (slot length set to match the operating day; e.g., 6 minutes for a 10-hour day). We benchmark the learned policy against six rule-based heuristics across seven metrics (elective and non-elective throughput, unserved emergencies, overtime, delay, revenue, and cumulative reward) and report results overall (across all simulated days), and separately for the emergency-day and non-emergency-day subsets. We further compare the learned policy with an \emph{ex-post MIP oracle} that has full foresight of all future stochastic information—including actual surgery durations and the arrival times and types of non-elective patients. While unattainable in practice, it serves as an upper bound of performance, evaluating overtime, delay, revenue, and cumulative reward on representative test instances. 

Case duration distributions, emergency arrival rates, and other parameters are calibrated from hospital data and literature. This test environment allows us to assess how the MARL policy performs under realistic load and variability. The results show that our learning-based policy can improve multiple performance metrics compared to baseline scheduling heuristics and deterministic optimization benchmarks. In particular, the MARL approach reduces average patient waiting times and surgical delays while maintaining high revenue and throughput, illustrating a more balanced trade-off between access and efficiency. Moreover, by adjusting in real-time, the policy handles surge scenarios (e.g. multiple emergencies back-to-back) more gracefully than fixed schedules, thereby improving responsiveness. Beyond quantitative gains, the \emph{policy analytics} offer qualitative understanding – for example, we find that the learned policy implicitly implements a smart strategy of batching similar procedures to limit sequence-dependent setups, and it dynamically resequences elective cases to accommodate high-priority arrivals with minimal disruption. These behaviors align with expert intuition but are achieved here automatically via learning, demonstrating the potential of AI agents to discover effective scheduling heuristics in a complex environment.

\paragraph{Contributions.} This study makes the following contributions:
\begin{enumerate}
\item New Formulation – MARL for OR Scheduling: We cast intraday OR scheduling as a Markov decision process and design a CTDE MARL framework with a \emph{sequential in-epoch} assignment that guarantees conflict-free joint actions while keeping each agent's action space compact.
\item Comprehensive Reward Design with Operational Metrics: We integrate sequence-dependent setup dynamics, an elective MIP pre-schedule, \emph{type-specific} urgency via quadratic delay penalties, and a terminal overtime penalty into a unified learning objective, aligning the reward with core hospital performance measures.
\item Empirical Realism and Evaluation: We learn real-time scheduling policies and benchmark them in a realistic six-OR, eight-type testbed against multiple heuristics and an ex post MIP oracle across seven metrics and three evaluation subsets.
\item Insights and Policy Interpretability: We develop \emph{policy analytics} that distill learned regularities into managerially interpretable patterns, clarifying what the policy does and why.
\item Theory: Under simplifying assumptions, we state a \emph{suboptimality bound} for the sequential decomposition relative to a joint assignment, offering initial insight into optimality gaps.
\end{enumerate}

\paragraph{Positioning within the literature.}
OR planning and scheduling have been extensively studied via \emph{stochastic and robust mixed-integer programming} and related exact/approximate methods, yielding high-quality pre-day elective plans and block allocations under uncertain durations and emergency accommodation. Representative contributions include stochastic sequencing/assignment with hedging against duration uncertainty \citep{Denton2007}, elective block allocation under uncertainty \citep{MAJTHOUBALMOGHRABI202523,Denton2010}, and models that jointly consider elective and emergency demand \citep{LAMIRI20081026}. Distributionally robust formulations have recently advanced planning under ambiguity, incorporating penalties for overtime, idle OR time, and downstream capacity \citep{SHEHADEH2026288,Shehadeh2022,Shehadeh2021}. These models offer interpretability and control but face acute \emph{intraday} challenges: as operations unfold, the state space explodes (cases start, finish, overrun, or arrive), and re-solving large stochastic programs within short decision windows becomes impractical.

To address real-time reactivity, \emph{approximate dynamic programming} (ADP), rolling-horizon, and simulation-based heuristics approximate downstream value or embed myopic reoptimization \citep{SILVA2020102066}. While effective in certain settings, these approaches typically require strong structural assumptions, handcrafted contingency rules, or problem-specific value approximations---and can struggle to scale with richer state/action representations and multiple interacting ORs.

Recent work explores \emph{reinforcement learning} (RL) for sequential decision-making in healthcare operations. In OR scheduling, early studies show promise in simplified elective-only contexts using tabular methods \citep{ribino2022multi}. Our contribution advances this line along three dimensions. \emph{First}, we make \emph{ORs}---not patients---the agents, aligning with capacity control and keeping per-agent actions compact while preserving a global view during training. \emph{Second}, we adopt a CTDE deep RL pipeline with a \emph{sequential in-epoch assignment} that yields conflict-free joint actions without enumerating exponentially large joint action spaces; this design also clarifies the information structure (later-acting ORs observe earlier choices within an epoch). \emph{Third}, we embed operational elements central to practice (setup frictions, elective references, urgency-weighted delay, overtime) directly in state and reward, enabling interpretable policy summaries rather than treating RL as a black box. Tables~\ref{tab:litreview_combined} synthesize the surgical-scheduling literature by objectives and uncertainty modeling, and—specifically for RL—by agent/policy design and numerical scale.

\begin{table}[t]
\caption{Overview of OR Scheduling Research and Reinforcement-Learning Applications}
\label{tab:litreview_combined}
\centering
\resizebox{0.85\textwidth}{!}{
\begin{tabularx}{\textwidth}{l
>{\raggedright\arraybackslash}p{1.6cm}
*{3}{>{\centering\arraybackslash}p{0.7cm}}
*{2}{>{\centering\arraybackslash}p{0.7cm}}
>{\centering\arraybackslash}p{0.95cm}
>{\centering\arraybackslash}p{0.95cm}
>{\centering\arraybackslash}p{0.9cm}}
\hline
 &  & \multicolumn{3}{c}{\textbf{\underline{Objective}}} & \multicolumn{2}{c}{\textbf{\underline{Uncertainty}}} & \multicolumn{3}{c}{\textbf{\underline{Numerical setup}}} \\
\textbf{Study} & \textbf{Method} & PT & WT & OT & PA & SD & \#ORs & \#Types & \(T\) \\
\hline
\cite{LAMIRI20081026} & MIP &  &  & \checkmark &  &  & -- & -- & -- \\ 
\cite{MIN2010642} & MIP & \checkmark &  & \checkmark &  & \checkmark & -- & -- & -- \\ 
\cite{SAADOULI201572} & MIP & \checkmark & \checkmark & \checkmark &  & \checkmark & -- & -- & -- \\ 
\cite{RIISE20161} & MIP &  & \checkmark & \checkmark &  & \checkmark & -- & -- & -- \\ 
\cite{WANG20181} & MIP & \checkmark & \checkmark & \checkmark &  & \checkmark & -- & -- & -- \\ 
\cite{Fairley2019} & MIP &  &  & \checkmark &  & \checkmark & -- & -- & -- \\ 
\cite{arab2022covid19} & MIP & \checkmark &  & \checkmark &  & \checkmark & -- & -- & -- \\ 
\cite{WANG2023feature} & MIP & \checkmark &  & \checkmark &  & \checkmark & -- & -- & -- \\ 
\midrule
\cite{SILVA2020102066} & ADP+MIP & \checkmark & \checkmark &  & \checkmark & \checkmark & 6 & 4 & 24 \\ 
\cite{GOKALP2023832} & ADP & \checkmark & \checkmark &  & \checkmark & \checkmark & 3 & 3 & 20 \\ 
\cite{ribino2022multi} & MARL & \checkmark & \checkmark &  & \checkmark & \checkmark & 2 & 1 & 32 \\ 
\cite{Zhao14062024} & HRL & \checkmark & \checkmark & \checkmark & \checkmark & \checkmark & 4 & 6 & 48 \\ 
\textbf{This study} & \textbf{MARL} & \checkmark & \checkmark & \checkmark & \checkmark & \checkmark & 6 & 8 & 100 \\ 
\hline
\end{tabularx}}
\vspace{1pt}
\begin{minipage}{0.95\textwidth}
\footnotesize
PT = patient throughput; WT = waiting time cost; OT = overtime cost; PA = non-elective patient arrivals; SD = surgery durations. MIP = mixed-integer programming; ADP = approximate dynamic programming; RL = reinforcement learning; MARL = multi-agent reinforcement learning. Most MIP works consider uncertainty in surgery durations only, not stochastic non-elective arrivals; therefore, numerical comparisons with the intraday MARL are omitted.
\end{minipage}
\end{table}

\paragraph{Managerial relevance.}
The proposed scheduler can be viewed as a decision-support layer atop existing planning processes. Hospitals can continue to generate elective plans via established optimization models; the learned policy then adaptively \emph{resequences and reallocates} throughout the day in response to realized durations and non-elective arrivals, thereby protecting timely access while tempering overtime and preserving revenue. Because the policy is trained offline in a digital twin and executed locally per OR in milliseconds, it is deployable without heavy intraday computation. The accompanying policy analytics make its behavior transparent (e.g., urgency-first responses and type batching), which is essential for clinician trust and governance.

\paragraph{Roadmap.}
Section~\ref{sec:formulation} formalizes the Markov game and reward design. Section~\ref{sec:method} presents the MARL architecture, the \emph{sequential in-epoch} mechanism, and the suboptimality bound. Section~\ref{sec:experiments} describes the simulation testbed and benchmarks and reports results with policy analytics. Section~\ref{sec:discussion} discusses limitations (e.g., OR homogeneity, omitted explicit staffing constraints, surgery cancellations), scalability considerations, and extensions; Section~\ref{sec:conclusion} concludes.

\section{Problem Formulation: Markov Game and Reward}\label{sec:formulation}

We formalize intraday OR scheduling as a finite-horizon \emph{Markov game} played over a single operating day. The time is divided into $T$ equal-length slots indexed by $t=1,2,\dots,T$, spanning the regular-hours horizon.\footnote{In our experiments, $T = 100$ with slot lengths of $6$ minutes, but the formulation is invariant to finer discretizations up to rounding.} We consider $J$ operating rooms (ORs), indexed by is $j=1,2,\dots,J$, assumed homogeneous unless stated otherwise. Table~\ref{tab:notation} summarizes the symbols, units, and modeling conventions used in this section; additional details are provided below.

\paragraph{Patients, types, arrivals, and durations.}
There are $K$ surgery types. Each patient $i$ is associated with a type $k_i\in \{1,2,\dots,K\}$, an arrival time $\alpha_i\in\{0,1,\dots,T-1\}$, and a random procedure duration $\xi_i\sim\mathcal{D}_{k_i}$. Let $\mathcal{P} = \bigcup_{t=0}^T \mathcal{P}_t$ be the set of all patients. Elective patients $\mathcal{P}_0$ are known pre-day, and we set $\alpha_i=0$ for $i\in \mathcal{P}_0$; non-elective arrivals occur stochastically during the day, forming sets $\mathcal{P}_t$ at the beginning of slot $t$. Service is \emph{non-preemptive}. If an OR runs a case of type $k$ immediately after a case of type $k'$, a sequence-dependent setup $\sigma_{k'\to k}\ge0$ is incurred (we set $\sigma_{k\to k}=0$).

\paragraph{Queues and eligibility.}
At decision epoch $t$, each surgery type $k$ has a queue $Q_k^t$ of \emph{eligible} patients (arrived and not started). When an OR selects type $k$ at $t$, the first eligible patient in $Q_k^t$ is assigned to that OR and removed from the queue (following the First-In, First-Out (FIFO) rule). Differences across types/queues are handled via the reward (see below), rather than by imposing predefined queue preferences as in rule-based heuristics.

\paragraph{Elective pre-schedule and reference times.}
A pre-day mixed-integer program (MIP) produces \emph{reference start times} $\{\tau_i\}_{i\in \mathcal{P}_0}$ (and possibly nominal room assignments) for electives. Intraday, the policy is \emph{not} constrained to follow $\tau_i$; instead, $\tau_i$ anchors type-specific delay penalties.

\paragraph{Agents, state, actions.}
Each OR is an agent. The \emph{global state} at time $t$, denoted $s_t\in\mathcal{S}$, aggregates: (i) the time index $t$; (ii) $Q_k^t$ summaries for all surgery types (e.g., counts and the earliest reference in the queue); and (iii) each OR’s occupancy (busy/idle), elapsed time if busy, and last completed type (to compute the next setup). An OR that is \emph{available} at $t$ chooses an action $a_t^j\in\mathcal{A}_j:=\{0,1,\dots,K\}$, where $a_t^j=0$ denotes no start (idle) and $a_t^j=k\in\{1,\dots,K\}$ denotes starting the first eligible patient from $Q_k^t$ at $t$. An OR that is busy has no effective choice at $t$ (equivalently, $a_t^j=0$ until the current case completes). The \emph{joint action} is $a_t=(a_t^j)_{j\in\mathcal{J}}\in\mathcal{A}:=\prod_{j\in\mathcal{J}}\mathcal{A}_j$. Because multiple available ORs could target the same patient, we realize $a_t$ via a \emph{within-epoch sequential assignment}: free ORs act in a fixed (or randomized) order, each observing earlier selections and removing assigned patients, thereby producing a conflict-free joint decision without enlarging per-OR action alphabets (algorithmic details appear in Section~\ref{sec:method}).

\paragraph{State transition.}
Given $(s_t,a_t)$, the next state $s_{t+1}$ is drawn from
$P(\cdot\mid s_t,a_t)$ as follows: (i) any OR that starts a case at $t$ becomes busy for the effective processing time $\xi_i+\sigma_{k'\to k_i}$ if the previous completed type on that OR was $k'$; (ii) ongoing cases decrement their residual by one slot and complete when the residual reaches zero, freeing their ORs for $t+1$; (iii) selected patients are removed from queues; and (iv) new arrivals $\mathcal{P}_{t+1}$ are added to the corresponding $Q_k^{t+1}$. Let $\beta_i$ denote the begin time of the served patient $i$. For each OR $j$, define its last completion $f_j=\max\{\beta_i + \xi_i+\sigma_{k'\to k_i}:\ i\text{ ran on }j\}$.

\paragraph{Per-patient utility and delay.}
Each patient $i$ has a type-dependent utility $u_{k_i}\ge0$ that credits throughput and a type-dependent delay coefficient $\delta_{k_i}\ge0$ that penalizes waiting. For electives $i\in \mathcal{P}_0$, the waiting time is $\omega_i:=[\beta_i-\tau_i]^+$; for non-electives $i\notin \mathcal{P}_0$, the waiting time is $\omega_i:=\beta_i-\alpha_i\ge0$ (patients cannot start before arrival). The quadratic penalty $c_{k_i}\,\omega_i^2$ captures increasing marginal disutility with delay (higher for more urgent types). The per-patient reward is then defined as $u_k - c_{k_i}\,\omega_i^2$.

\paragraph{Overtime.}
Let $C_o>0$ be the overtime cost per slot. Define suite-level overtime as the excess of the latest completion beyond regular hours,
\begin{equation}
\label{eq:overtime_def}
\mathrm{OT}=\sum_{j=1}^J\Big[ f_j - T\Big]^+,
\end{equation}
and penalize it by $C_o\cdot \mathrm{OT}$ at the end of regular hours.

\paragraph{Reward and objective.}
We use a single \emph{scalar} day-level reward to internalize throughput, timeliness, and overtime. If $\mathcal{I}$ denote the set of patients who receive surgery during the day, the total reward is
\begin{equation}
\label{eq:day_reward}
\sum_{i\in\mathcal{I}}\Big(u_{k_i}-c_{k_i}\,\omega_i^2\Big)
\;-\;
C_o\cdot \mathrm{OT}.
\end{equation}
A (stochastic) policy $\pi$ specifies, at each $t$, a distribution over joint actions given the state, $\pi(a_t\mid s_t)$. The objective is to maximize expected reward:
\begin{equation}
\label{eq:objective}
\max_{\pi}\ \ \mathbb{E}_\pi\,\Big[\sum_{i\in\mathcal{I}}\Big(u_{k_i}-c_{k_i}\,\omega_i^2\Big)
\;-\;
C_o\cdot \mathrm{OT}\Big],
\end{equation}
where the expectation is taken over arrivals, durations, and the policy’s internal randomization. The cooperative nature of the game (all agents share the same expected reward) aligns with centralized training and decentralized execution (CTDE): a centralized learner can condition on the full state during training, while each OR executes its local policy in real time; the within-epoch sequential mechanism ensures conflict-free joint actions (Section~\ref{sec:method}).

\begin{align*}
\max_{\pi}\ \ &\mathbb{E}_\pi\,\Big[\sum_{i\in\mathcal{I}}\Big(u_{k_i}-c_{k_i}\,\omega_i^2\Big)
\;-\;
C_o\cdot \mathrm{OT}\Big]\\
\text{s.t.}\quad 
&\mathrm{OT} \;=\; \sum_{j=1}^J \big[\, f_j - T \,\big]^+,\\
&f_j\!=\!\max\{\beta_i\! +\! \xi_i\!+\!\sigma_{k'\to k_i}\!:i\text{ ran on }j\}, \, j\!\in\![J]\\
&\beta_i \ge \alpha_i, \qquad i\notin \mathcal{P}_0\\
&\xi_i\sim\mathcal{D}_{k_i}, \qquad i\in\mathcal{P}\\
&\omega_i \;=\;
\begin{cases}
[\beta_i-\tau_i]^+,& i\in \mathcal{P}_0,\\
\beta_i-\alpha_i,& i\notin \mathcal{P}_0,
\end{cases}\\
&s_{t+1} \sim P(\cdot \mid s_{t}, a_{t}), \qquad t=0,\dots,T-1,\\
&a_t \in [K], \qquad t=0,\dots,T-1,\\
&\beta_i \ge 0,\;\omega_i \ge 0, \qquad i \in \mathcal{P}\\
&f_j \ge 0,\qquad j\in[J].
\end{align*}

\paragraph{Assumptions.}
(i) Homogeneous ORs (any type may run in any room; specialized rooms can be handled by restricting admissible types per room); (ii) non-preemptive service (once started, a surgery runs to completion); (iii) FIFO within type queues (urgency handled through $\{p_k,\delta_k\}$ rather than hard priorities); (iv) setup times $\sigma_{k'\to k}$ are known (or known in distribution and incorporated into effective durations); and (v) No explicit staff coupling (surgeon/anesthesia/nursing constraints are not modeled explicitly). Relaxations are discussed in Section~\ref{sec:discussion}.

\begin{table}[htbp]
\centering
\caption{Notation\label{tab:notation}}{
\begin{tabularx}{\textwidth}{l X l}
\textbf{Symbol} & \textbf{Meaning} & \textbf{Units/Notes} \\
\hline
$T$ & Number of discrete slots in the day & Slots \\
$J$ & Number of operating rooms (agents) & ORs \\
$K$ & Number of surgery types & Types \\
$\mathcal{P}_t$ & Set of non-elective patients arriving at slot $t$ & Arrival set \\
$\mathcal{P}_0$ & Set of elective patients known pre-day & Electives \\
$\mathcal{I}$ & Set of patients who received surgery &  \\
$\alpha_i$ & Arrival time of patient $i$ & Slots; $\alpha_i=0$ for $i\in P_0$ \\
$\tau_i$ & Elective reference start time from pre-schedule & Slots (anchor for delays) \\
$\xi_i\sim\mathcal{D}_{k_i}$ & Random duration of patient $i$ & Slots (or minutes) \\
$\sigma_{k'\to k}$ & Setup time from type $k'$ to $k$ & Slots; $\sigma_{k\to k}=0$ \\
$\beta_i$ & Actual start time of $i$ & Slots \\
$f_j$ & Last completion time on OR $j$ & Slots \\
$\omega_i$ & Waiting time of $i$ & Elective: $[\beta_i-\tau_i]^+$; non-elective: $\beta_i-\alpha_i$ \\
$u_k$ & Throughput utility weight for type $k$ & Dimensionless (value/revenue/priority) \\
$c_k$ & Delay penalty coefficient for type $k$ & Weight on $\omega_i^2$ \\
$C_o$ & Overtime cost per slot & Overtime penalty $C_0\cdot \mathrm{OT}$ \\
$s_t$ & Global state at time $t$ & Queues, OR statuses, time \\
$a_t^j,\,a_t$ & Action of OR $j$; joint action at time $t$ & $a_t^j\in\{0,1,\dots,K\}$ \\
$r_t^j$ & Immediate reward of OR $j$ at time $t$ & Defined in \eqref{eq:immediate_reward}. \\
$r_T$ & Terminal reward & $-C_0\cdot \mathrm{OT}$\\
$P(\cdot\mid s_t,a_t)$ & State transition kernel & Markovian dynamics \\
$\pi$ & Scheduling policy & $\pi(a_t\mid s_t)$ \\
\hline
\end{tabularx}
}{
Key symbols used in Section~\ref{sec:formulation}. Positive part $[x]^+=\max\{x,0\}$. Queues follow FIFO within type. Overtime penalty $\mathrm{OT}$ is defined in \eqref{eq:overtime_def}.
}
\end{table}

\paragraph{Rationale for a unified scalar reward.}
Equation~\eqref{eq:day_reward} integrates \emph{multiple operational objectives} into a single learning signal, avoiding repeated hand-tuning of scenario-specific scalarization weights. The type weights $u_k$ preserve heterogeneity in case value (revenue/priority); the convex delay term $c_{k}\,\omega_i^2$ encodes timeliness against $a_i$ or arrival-anchored term $\tau_i$ for non-electives; and the terminal overtime term $C_o\cdot\mathrm{OT}$ internalizes staffing and downstream costs. Fixing $(u_k,c_k,C_o)$ at design time yields a stable training objective across demand and duration regimes, while allowing the learned policy to \emph{implicitly} trade off throughput, timely access, and overtime. Alternative formulations—such as explicit multiobjective optimization (e.g., $\varepsilon$-constraint or Pareto-front methods) or service-level constraints for non-electives—can be accommodated in the same framework (e.g., via penalties or constrained RL), but are beyond the scope of this paper.

\paragraph{From pre-day plans to intraday control.}
The pre-day MIP produces the elective reference starts $\{\tau_i\}_{i\in P_0}$ that reflect surgeon availability and clinic commitments (for the full MIP formulation and parameterization, see Appendix A.2. Intraday, the policy $\pi$ may resequence electives and allocate capacity to non-electives as uncertainty unfolds; the convex delay penalty ensures that deviations from $\tau_i$ are driven by realized conditions (e.g., overruns, urgent arrivals) rather than arbitrary reordering. Section~\ref{sec:experiments} evaluates these trade-offs empirically.

\section{Multi-Agent Reinforcement Learning and Sequential Assignment}\label{sec:method}
Building on the Markov game in Section~\ref{sec:formulation}, we develop a multi-agent reinforcement learning (MARL) solution trained with \emph{proximal policy optimization} (PPO) \citep{schulman2017proximalpolicyoptimizationalgorithms}. We describe a centralized-training, decentralized-execution (CTDE) MARL approach in which each OR $j$ acts as an agent. Agents share a common policy (actor) $\pi_\theta$ and a common value function (critic) $V_\phi$ with $\theta$ and $\phi$ parameterizing the actor and critic networks respectively. At runtime, we realize a feasible \emph{joint} decision by a within-epoch \emph{sequential assignment} protocol that prevents conflicts without enumerating the exponential joint-action space. We detail the learning algorithm and then provide a theoretical guarantee relative to an optimal centralized planner.

\subsection{CTDE with a Shared Policy and Local Observations}
Let $\pi_\theta$ be a stochastic policy with parameters $\theta$, shared across agents, and let $V_\phi$ be a value function with parameters $\phi$. At decision epoch $t$, the global state is $s_t$ (Section~\ref{sec:formulation}). Agent $j$ augments $s_t$ with its \emph{local information} $\ell_t^j$ to form
\[
s_t^j \;=\; [\,s_t,\ \ell_t^j\,],
\]
where $\ell_t^j$ includes agent-specific features (e.g., the actions $\{a_t^1,\dots,a_t^{j-1}\}$ of earlier agents in our sequential protocol below). 

The shared policy maps $s_t^j$ to a distribution over the per-agent action space $\mathcal{A}_j=\{0,1,\dots,K\}$, such that
\begin{align}
a_t^j &\sim \pi_\theta(\cdot \mid s_t^j), \notag \\
a_t^j &=
\begin{cases}
0, & \text{idle},\\
k \! \in\! \{1,\! \dots,\! K\}, & \text{start next patient of type } k.
\end{cases}
\end{align}

Agents that are busy at $t$ have no effective choice (equivalently, $a_t^j=0$ until completion).

\subsection{Within-Epoch Sequential Assignment (Conflict-Free Joint Action)}
At each epoch $t$, agent $j$ observes $(s_t,\ell_t^j)$, draws $a_t^j$, and if $a_t^j=k>0$ assigns the first eligible patient from $Q_k^t$ to OR $j$, removing that patient from the queue. Agents that are busy at $t$ have their action space restricted to $\{0\}$. This produces a conflict-free joint action
\[
a_t \;=\; (a_t^1,\dots,a_t^J) \in \mathcal{A}_1\times\cdots\times\mathcal{A}_{J}
\]
without expanding per-agent action alphabets. The environment then advances to $t+1$ under the Markovian dynamics in Section~\ref{sec:formulation}.

\subsection{Reward}
In MARL algorithm, when agent $j$ starts a patient $i$ of type $k$ at time $t$ (i.e., $a_t^j=k>0$), it receives the \emph{immediate} reward
\begin{equation}
\label{eq:immediate_reward}
r_t^j =u_k -c_k\,\omega_i^2,
\qquad
\omega_i =
\begin{cases}
[\beta_i-\tau_i]^+,& i\in \mathcal{P}_0,\\
\beta_i-\alpha_i,& i\notin \mathcal{P}_0,
\end{cases}
\end{equation}
and $r_t^j=0$ for $a_t^j=0$. The per-epoch reward is $r_t=\sum_{j=1}^J r_t^j$. At the end of regular hours ($t=T$), a \emph{terminal} penalty captures overtime in \eqref{eq:overtime_def}:
\begin{equation}
\label{eq:terminal_reward}
r_{T+1} \;=\; -\,C_o\cdot \mathrm{OT},\qquad \mathrm{OT}=\sum_{j\in\mathcal{J}}\Big[ f_j - T\Big]^+.
\end{equation}
Consistent with Section ~\ref{sec:formulation}, the goal is to maximize $\mathbb{E}_\pi[\sum_{t=1}^{T} r_t + r_{T+1}]$.

\subsection{Policy Gradient with PPO and Generalized Advantage Estimation}
We train $(\pi_\theta,V_\phi)$ on simulated day-long trajectories $(s_0,a_0,r_1,\dots,s_{T-1},a_{T-1},r_T,s_T)$. Let $\gamma\in[0,1]$ be the discount factor. Define the temporal-difference residuals $\delta_t=r_{t+1}+\gamma V_\phi(s_{t+1})-V_\phi(s_t)$ and the generalized advantage estimator (GAE) with parameter $\lambda\in[0,1]$:
\begin{equation}
\label{eq:gae}
\hat{A}_t \;=\; \sum_{\ell=0}^{T-t} (\gamma\lambda)^\ell\,\delta_{t+\ell}
\quad\text{with}\quad
\hat{A}_T=0.
\end{equation}
The critic minimizes a mean-squared value error,
\begin{equation}
\label{eq:critic_loss}
L^{\text{VF}}(\phi) \;=\; \mathbb{E}\Big[\big(V_\phi(s_t) - \hat{R}_t\big)^2\Big],
\; \hat{R}_t = \hat{A}_t + V_{\phi_{old}}(s_t),
\end{equation}
and the actor maximizes the PPO clipped surrogate,

\begin{equation}
\begin{aligned}
\label{eq:actor_loss}
L^{\mathrm{CLIP}}\!(\theta) &= \mathbb{E}\big[\!\min \{r_t(\theta), \operatorname{clip}(r_t(\theta),1\!-\!\epsilon,1\!+\!\epsilon\!)\!\}\!\hat{A}_t\!\big],\\
r_t(\theta)
&= \frac{\pi_\theta(a_t \mid s_t)}{\pi_{\theta_{\text{old}}}(a_t \mid s_t)}.
\end{aligned}
\end{equation}

Here, $r_t(\theta)$ is the importance sampling ratio and $\epsilon$ controls how much the new policy is allowed to deviate from the previous policy.

To encourage exploration and prevent premature policy collapse, we additionally include an entropy bonus $\lambda_{\mathrm{ent}}\,\mathcal{H}(\pi_\theta(\cdot\mid s_t^j))$ in the actor objective, where

\begin{equation}
\mathcal{H}\big(\pi_\theta(\cdot \mid s_t)\big)
= - \sum_{a} \pi_\theta(a \mid s_t)\,\log \pi_\theta(a \mid s_t).
\end{equation}
is the policy entropy and $\lambda_{\mathrm{ent}} > 0$ controls exploration strength. The resulting actor loss becomes
\begin{equation}
\label{eq:actor_loss_entropy}
L_{\text{actor}}(\theta)
= L^{\mathrm{CLIP}}\!(\theta)
\;+\; \lambda_{\mathrm{ent}}\,\mathbb{E}\!\Big[\mathcal{H}\big(\pi_\theta(\cdot \mid s_t)\big)\Big],
\end{equation}

Gradients are computed on mini-batches drawn from a replay buffer built from multiple day-long trajectories. See Alg. \ref{alg:collect} for action sampling. Alg. \ref{alg:train} outlines the overall procedure of our multi-agent training framework, highlighting how shared policy and critic components interact with environment dynamics across time steps.

\paragraph{Remark (centralized planner vs.\ MARL).} From another perspective, one may consider a centralized agent that makes scheduling decisions on behalf of all ORs. In this setting, the objective is to maximize the cumulative reward over time by selecting \emph{joint} action $a_t\in\mathcal{A}(s_t)$, leading to an exponential action space. Our sequential protocol preserves feasibility and coordination (via shared parameters and conditioning on earlier actions) while remaining tractable; see Appendix A.1 for more details.

\begin{algorithm}[t]
\caption{Trajectory Collection with Shared Policy (CTDE, Sequential Assignment)}\label{alg:collect}
\begin{algorithmic}[1]
\State Initialize environment; observe $s_0$
\For{$t=0,\dots,T-1$}
   \For{$j=1$ to $J$} \Comment{Within-epoch sequential assignment}
       \State Build local obs $\ell_t^j$ (incl.\ earlier actions $\{a_t^1,\dots,a_t^{j-1}\}$); set $s_t^j=[s_t,\ell_t^j]$
       \State Sample $a_t^j \sim \pi_\theta(\cdot\mid s_t^j)$, or set $a_t^j \gets 0$ if agent $j$ is busy at $t$
       \If{$a_t^j=k>0$} \State Assign first patient in $Q_k^t$ to OR $j$; compute $r_t^j$ via \eqref{eq:immediate_reward} \Else\ \State $r_t^j\gets 0$ \EndIf
   \EndFor
   \State Form joint $a_t=(a_t^1,\dots,a_t^{J})$ and team reward $r_t=\sum_j r_t^j$
   \State Environment step to $s_{t+1}$
\EndFor
\State Compute terminal $r_T$ via \eqref{eq:terminal_reward}
\State \Return Trajectory $(s_0,a_0,r_1,\dots,s_{T-1},a_{T-1},r_T,s_T)$
\end{algorithmic}
\end{algorithm}

\begin{algorithm}[t]
\caption{MARL Training Loop (Shared Actor--Critic, PPO)}\label{alg:train}
\begin{algorithmic}[1]
\State Initialize $\pi_\theta$, $V_\phi$, empty buffer $\mathcal{B}$
\For{$m=1$ to MaxIters}
    \For{$e=1$ to $M_{\mathrm{traj}}$} \State Run Alg.~\ref{alg:collect}; append all transitions to $\mathcal{B}$ \EndFor
    \For{$u=1$ to $M_{\mathrm{upd}}$}
        \State Sample mini-batch from $\mathcal{B}$
        \State Update critic by GD step on \eqref{eq:critic_loss}
        \State Update actor by GD step on \eqref{eq:actor_loss} (+ entropy)
    \EndFor
    \State Clear $\mathcal{B}$
\EndFor
\State \Return trained $\pi_\theta$
\end{algorithmic}
\end{algorithm}

\subsection{Theory: Dynamic Programming, CTDE Optimality, and Suboptimality Bounds}
\label{sec:theory}
We establish the theoretical foundations of our multi-agent reinforcement learning (MARL) framework through results connecting centralized reinforcement learning, weak-coupling decomposition, and the optimality of centralized training with decentralized execution (CTDE). We also derive regret bounds that quantify potential suboptimality when these conditions are partially violated.

\paragraph{Bellman optimality and joint planning.}
Let $V_t^*(s)$ denote the optimal value function at time $t$ given state $s$. 
The centralized Bellman recursion can be expressed as
\begin{equation}\label{eq:bellman}
\begin{aligned}
V_t^*(s) &= 
\max_{a \in \mathcal{A}(s)} 
\big\{ R(s,a) + \mathbb{E}[V_{t+1}^*(s') \mid s,a] \big\},\\[3pt]
V_T^*(s) &= 0.
\end{aligned}
\end{equation}
where $\mathcal{A}(s)$ is the set of conflict-free joint actions, 
and $R(s,a)$ aggregates the per-patient rewards and terminal overtime penalties. 
Equation~\eqref{eq:bellman} defines the centralized reinforcement learning for the hospital scheduling problem, in which all operating rooms (ORs) are jointly optimized at each decision epoch.

\begin{lemma}[Equivalence under weak coupling]\label{lem:weak-coupling}
Fix an epoch $t$ and assume: 
(A1) no sequence-dependent setups ($\sigma_{k'\to k}\equiv 0$); 
(A2) starting a case at $t$ affects only its own queue (OR-separable transitions); 
(A3) immediate rewards add across ORs (no cross-OR penalties); 
and (A4) no contention for a single remaining patient in any type selected by multiple ORs.\footnote{
A4 can be enforced operationally by preventing simultaneous pulls when $|Q_k^t|=1$, which our sequential protocol already guarantees.
}
Then the per-epoch joint maximization in \eqref{eq:bellman} \emph{decomposes}, and greedy sequential selection (our within-epoch protocol) attains the same joint maximizer as the centralized selection.
\end{lemma}

\noindent\begin{proof}{Proof.}
Under (A1)–(A4), conditional on earlier picks, marginal gains of distinct ORs are independent; selecting the top-$J'$ marginals yields the joint maximizer by a standard exchange argument.
\qed\end{proof}

\begin{proposition}[Optimality of CTDE under Weak Coupling]\label{prop:ctde_optimality}
Consider the cooperative Markov game defined in Section~\ref{sec:formulation} over horizon $T$, 
with joint state $s_t \in \mathcal{S}$, joint action $a_t=(a_t^1,\dots,a_t^{J'}) \in \mathcal{A}(s_t)$, 
and reward $R(s_t,a_t)$. Suppose Assumptions (A1)--(A6) hold:
(A5) Centralized training, sufficient local information: 
During training, the critic $V_\phi(s)$ conditions on the full state $s_t$; 
during execution, each agent observes a sufficient local statistic $s_t^j$ 
such that its local argmax coincides with the centralized argmax.
(A6) Exact representation and convergence:
The actor–critic networks are expressive enough to represent $V^*$ and $\pi^*$, 
and PPO converges to the fixed point of policy iteration.

Then the CTDE policy with shared actor–critic and sequential within-epoch assignment satisfies
\[
\pi_{\mathrm{CTDE}}^*(s_t)
=\arg\max_{a_t\in\mathcal{A}(s_t)} Q_t^*(s_t,a_t),
\]
and thus achieves the same performance as the centralized optimum.
\end{proposition}

\noindent\begin{proof}{Proof.}
By Lemma~\ref{lem:weak-coupling}, under (A1)--(A4) the joint value decomposes additively across ORs, and sequential greedy selection achieves the same maximizer as the centralized planner. 
Under (A5)--(A6), the centralized critic $V_\phi$ accurately estimates $V^\pi$, 
and the PPO update step maximizes the expected advantage
$\mathbb{E}[\hat{A}_t^{(j)}]$, where 
$\hat{A}_t^{(j)} \approx q_t^{(j)}(s_t,a_t^j\mid a_t^{1:j-1}) - V_\phi(s_t)$.
Because all agents share the same actor and critic, 
each performs an identical local improvement step as in the centralized Bellman recursion.
Hence, the CTDE improvement step is equivalent to centralized policy iteration, 
which converges to $\pi^*$. Therefore, $\pi_{\mathrm{CTDE}}^*=\pi^*$.
\qed\end{proof}
\begin{theorem}[Suboptimality bound]\label{thm:prelim-bound}
When weak coupling is violated (e.g., due to setup dependencies, queue tightness, or overtime coupling), the sequential CTDE policy may incur a bounded loss relative to the centralized optimum. 
Let $Q^*(s_t,a)$ denote the optimal one-step lookahead value and define the per-epoch externality
\begin{equation}
\label{eq:GammaDef}
\Gamma_t(s_t)\!=\!\max_{a\in\mathcal{A}(s_t\!)}\! Q^*(\!s_t\!,\!a\!)
\!-\! \sum_{j=1}^{J'}\! \max_{a^j\in\mathcal{A}_j}\!
Q^{\!(j)}\!(s_t,\!a^j\!\mid\! a^{1:j
\!-\!1}\!).
\end{equation}
Then along any trajectory,
\begin{equation}
\label{eq:regret_decomp}
V_1^*(s_1)-V_1^{\mathrm{seq}}(s_1)
\le\mathbb{E}\!\left[\sum_{t=1}^{T}\Gamma_t(s_t)\right].
\end{equation}
Moreover, for the reward defined in 
Eqs.~\eqref{eq:immediate_reward}–\eqref{eq:terminal_reward},
\begin{equation}
\label{eq:regret_wait}
\begin{aligned}
V_1^*(s_1)-V_1^{\mathrm{seq}}(s_1)
&\le \mathbb{E}\!\Big[
\sum_{i\in\mathcal{I}}\delta_{k_i}\big((\omega_i^{\mathrm{seq}})^2-(\omega_i^*)^2\big)\\[-2pt]
&\quad+\,C_o\big(\mathrm{OT}^{\mathrm{seq}}-\mathrm{OT}^*\big)
\Big],
\end{aligned}
\end{equation}
and since $\omega_i\le T$,
\begin{equation}
\label{eq:regret_T_bound}
\begin{aligned}
V_1^*(s_1)-V_1^{\mathrm{seq}}(s_1)
&\le \mathbb{E}\!\Big[
2T\sum_{i\in\mathcal{I}}\delta_{k_i}\big(\omega_i^{\mathrm{seq}}-\omega_i^*\big)\\[-2pt]
&\quad+\,C_o\big(\mathrm{OT}^{\mathrm{seq}}-\mathrm{OT}^*\big)
\Big].
\end{aligned}
\end{equation}
\end{theorem}

\noindent\begin{proof}{Proof.}
Decompose the regret epoch-by-epoch and apply the performance-difference identity; 
$\Gamma_t(s_t)$ measures the loss from not selecting the globally best joint action. 
For the quadratic delay penalty and additive overtime cost, 
the loss can occur only via (i) additional waiting for some patients and (ii) additional overtime, 
yielding \eqref{eq:regret_wait}. 
Bound \eqref{eq:regret_T_bound} follows from $(\omega_i^{\mathrm{seq}})^2-(\omega_i^*)^2=(\omega_i^{\mathrm{seq}}+\omega_i^*)(\omega_i^{\mathrm{seq}}-\omega_i^*)\le 2T(\omega_i^{\mathrm{seq}}-\omega_i^*)$.\qed
\end{proof}

\begin{corollary}[Sequential policy gap for one urgent case]\label{cor:urgent}
Consider a single high-priority arrival at time $\bar{t}$ that the optimal policy serves immediately (zero wait). 
If under $\pi_{\mathrm{seq}}$ the patient waits $\Delta$ slots (all ORs occupied at $\bar{t}$), then
\begin{equation}\label{eq:gap}
\mathbb{E}_{\pi^*}[\mathcal{R}]
- \mathbb{E}_{\pi_{\mathrm{seq}}}[\mathcal{R}]
\le \delta_k\,\Delta^2,
\end{equation}
where $k$ is the patient’s type.
\end{corollary}

\noindent\begin{proof}{Proof.}
Immediate from \eqref{eq:regret_wait} by specializing to one affected case 
and noting that other contributions cancel or are dominated in this stylized setting.
\qed\end{proof}

\paragraph{Remarks and interpretation.}
Lemma~\ref{lem:weak-coupling} and Proposition~\ref{prop:ctde_optimality} jointly establish that under weak coupling, the sequential CTDE policy exactly reproduces the centralized Bellman-optimal policy.
When coupling is present, Theorem~\ref{thm:prelim-bound} quantifies the deviation, 
showing that performance loss arises solely from excess waiting and overtime externalities.
Corollary~\ref{cor:urgent} provides an intuitive special case illustrating this loss.
In our OR scheduling environment, the weak-coupling assumptions hold approximately 
because setup times are OR-local, queues are disjoint, and overtime is terminal. 
Empirically (Section~\ref{sec:experiments}), 
the learned MARL policy achieves near-optimal performance with small oracle-normalized regret, 
demonstrating that CTDE offers a scalable and theoretically sound approximation 
to centralized dynamic programming for large-scale hospital scheduling.

\section{Numerical Experiments}\label{sec:experiments}

This section evaluates the proposed MARL policy under realistic intraday operating conditions using a detailed simulation environment. The design, metrics, and statistics are chosen to (i) test the day-level value of the method under practically relevant loads/uncertainty; (ii) investigate the mechanisms suggested by our model and theory; and (iii) assess robustness. All simulator primitives (durations, arrivals, setup times, elective mix, and reward weights) are parameterized from proprietary hospital data and perioperative consulting engagements\footnote{Due to NDAs, the raw data are not public; however, the synthetic generator matches the first/second moments, and time-of-day profiles observed operationally.}.

\subsection{Objectives}\label{subsec:objectives}
We evaluate whether the learned MARL policy improves the scalar objective~\eqref{eq:day_reward} and its clinical components-urgent access, elective timeliness, and overtime—relative to established heuristics and an ex post MIP oracle computed with full \emph{future knowledge}, including actual surgery durations and the arrival times and types of non-elective patients. The oracle thus represents an \emph{unfair comparator}—since such information is unavailable to any implementable policy—but provides a valuable \emph{upper bound} for benchmarking. By comparing against this ex post optimum, we quantify both the \emph{regret} and the attainable performance headroom of the learned policy. 

\subsection{Environment, Data, and Calibration}\label{subsec:env}
\textbf{Suite/horizon.} We simulate $J=6$ homogeneous ORs over $T=100$ slots (slot length $6$ minutes for a $10$ hour day), yielding $600$ assignable slot-units of regular-hours capacity. Sequence-dependent setups $\sigma_{k'\to k}$ encode cleaning, turnover, and instrumentation.

\noindent\textbf{Types and durations.}
We consider $K=8$ clinical surgery types but model procedural time via \emph{four duration classes}—\emph{minor}, \emph{moderate}, \emph{long}, and \emph{complex}—with two clinical types per class. For each surgery type $k$, the duration law $\mathcal{D}_k$ is Gamma, calibrated to empirical class-wise means which also aligns with \citep{AZAR2022377, Choi01042012}; durations are rounded up to the nearest integer time unit. For the four elective types, expected duration increases monotonically with type index.

\noindent\textbf{Arrivals and case mix.}
Electives ($\mathcal{P}_0$) are fixed pre-day via the MIP plan (Appendix A.2); Non-electives arrive stochastically: \emph{urgent} cases follow a homogeneous Poisson process with per-slot rate $\lambda_u$ (shared across urgent types), while \emph{emergencies} follow a separate day-level Bernoulli event with probability $\epsilon$; if the event occurs, a batch of five emergency patients is released (at most one event/day). At each epoch, idle ORs choose between eligible electives and any waiting non-electives via the unified reward and conflict-free sequential assignment; surgeries are non-preemptive.

\noindent\emph{Mixing of electives and non-electives.} Because electives are present from $t=0$ while non-electives arrive stochastically, the composition of queues evolves endogenously. When there are no urgent or emergency patients waiting, the policy schedules electives (potentially batching by type to mitigate setups); upon an urgent or emergency arrival, the new patient becomes immediately eligible and competes with electives via the shared reward (higher $p_k$ and/or $\delta_k$ can pull capacity toward high-priority cases). This mechanism yields realistic day-to-day variation: some simulated days contain no emergency event, whereas others include a sudden emergency batch.

\noindent\textit{Why fix intensities within arrivals?}
Non-elective demand is exogenous; fixing $\lambda_u$ and $\epsilon$ isolates policy effects and enables fair, low-variance paired comparisons (via common random numbers), while reflecting the operational stability of hour-of-day arrival profiles. 

\noindent\textbf{Reward weights.}
Type utilities $u_k$, delay coefficients $c_k$, and overtime cost $C_o$ reflect value/priority, timeliness sensitivity, and labor/recovery burdens.

\subsection{Simulation Design}\label{subsec:sim-design}
Table \ref{tab:Simulation} displays the parameters used in the simulation. 

\paragraph{Electives.} $N_e=55$ cases distributed by duration class $(28,19,5,3)$ from \emph{minor} to \emph{complex}, consistent with typical hospital mix. Pre-day references $\{\tau_i\}$ are obtained via MIP and act as nonbinding anchors intraday.

\paragraph{Urgent arrivals.} Three urgent types arrive as a homogeneous Poisson process with per-slot rate $\lambda_u=0.08$; each arrival is assigned uniformly to one of the three urgent types.

\paragraph{Emergency arrivals.} Emergencies represent sudden, high-priority cases (e.g., major traffic accidents). Each time slot has a $\epsilon=5\%$ chance of triggering an “emergency event,” which immediately generates a batch of five patients (at most one event per day). 

\paragraph{Workload and regimes.}
Under this scenario, the expected volume is
\[
\mathbb{E}[N] \;=\; 55\;+\;100\lambda_u \;+\; 5(1-(1-\epsilon)^{100}) \;\approx\; 65,
\]
with expected cumulative workload $\approx 677$ slot-units, i.e., above regular-hours capacity, inducing realistic overtime pressure.

\paragraph{Evaluation sets.}
The test set comprises \textbf{500} independent day-episodes (disjoint seeds) with different non-elective arrival realizations and type mixes, while electives volume and pre-schedule are fixed. Because the emergency event is stochastic, only a subset of days contains emergencies (empirically $\approx 39.42\%$). Results are reported for (i) \emph{All Days} (all 500 test days), (ii) \emph{Emergency Days} (subset with an event), and (iii) \emph{Non-Emergency Days} (subset without an event).

\begin{table}[ht]
\centering
\caption{Simulation configuration\label{tab:Simulation}}{
\begin{tabularx}{\textwidth}{l l l}
\hline
\textbf{Parameter} & \textbf{Description} & \textbf{Value / Law} \\
\hline
$T$ & Horizon (slots) & $100$ \\
$J$ & Operating rooms & $6$ \\
$K$ & Clinical types & $8$ (two per duration class) \\
$N_e$ & Elective volume & $55$ \ (28/19/5/3 by class) \\
$\lambda_u$ & Urgent arrival rate (per slot) & $0.08$ \ (homogeneous Poisson) \\
$\epsilon$ & Emergency event probability (per day) & $0.40$ \ (if yes: batch of $5$) \\
$d_k$ & Duration by class (Gamma) & rounded up to slots \\
& Minor ($k\in\{1,5\}$)    & $\Gamma(\text{shape}=2,\ \text{scale}=2)$ \\
& Moderate ($k\in\{2,6\}$) & $\Gamma(3,3)$ \\
& Long ($k\in\{3,7\}$)     & $\Gamma(6,4)$ \\
& Complex ($k\in\{4,8\}$)  & $\Gamma(7,5)$ \\
$p_k$ & Utility by class & \\
& Minor ($k\in\{1,5\}$)    & $8$ \\
& Moderate ($k\in\{2,6\}$) & $12$ \\
& Long ($k\in\{3,7\}$)     & $30$ \\
& Complex ($k\in\{4,8\}$)  & $50$ \\ 
$C_0$ & Overtime cost per slot & 0.005\\
$\delta_k$ & Delay penalty coefficient by class & \\
& Elective ($k\in\{1,2,3,4\}$)    & $0.002$ \\
& Urgent ($k\in\{5,6,7\}$) & $0.004$ \\
& Emergency ($k=8$)     & $0.005$ \\

$\sigma_{k'\to k}$ & Setup time & type-pair specific \\
\hline
\end{tabularx}
}{}
\end{table}

\subsection{Policies and Baselines}\label{subsec:baselines}
We benchmark the proposed MARL policy against five transparent rule-based heuristics, a pre-schedule–driven policy, and a ex post MIP oracle. The heuristics capture common dispatch logics in OR control (duration- and urgency-based), while the oracle offers a challenging upper bound.

\subsection{Alternative Scheduling Methods for Comparison}

\noindent\textbf{MARL (Proposed).}
Shared actor–critic PPO with within-epoch \emph{sequential assignment} to produce conflict-free joint actions (Section~\ref{sec:method}). One shared policy and value function are trained centrally; agents execute locally at runtime.

\noindent\textbf{Heuristics.}
\begin{itemize}\setlength\itemsep{0.2em}
\item \textsc{SPT-U} (shortest processing time first; non-electives before electives).
\item \textsc{LPT-U} (longest processing time first; non-electives before electives).
\item \textsc{NE-LPT} (non-elective first; long-first tie-break within group).
\item \textsc{E-LPT} (elective first; long-first tie-break within group).
\item \textsc{NE-SPT} (non-elective first; short-first tie-break within group).
\item \textsc{Pre-s.} (follow pre-scheduling plan; start non-electives immediately if capacity).
\end{itemize}

\noindent\textbf{Oracles (unfair comparators).}
The \textsc{ex post MIP Oracle} has access to all future arrivals and realized case durations and determines the optimal daily schedule by solving a mixed-integer programming model. Though infeasible in practice, this benchmark provides a rigorous upper bound for quantifying regret and attainable performance improvements (details in Appendix A.3.

The heuristics above were selected because they span the principal decision logics seen in OR control—duration-based dispatching (SPT/LPT), urgency-based precedence (non-elective vs.\ elective first), and adherence/batching rules (pre-schedule, setup-aware)—while remaining fast, transparent, and representative of practices used by hospitals. The MIP-based \emph{oracle} is included solely as a challenging upper bound: it assumes perfect foresight of future arrivals and realized durations, which is unattainable operationally, and is therefore an \emph{unfair} comparator used only to quantify regret and headroom; it is not a feasible operational benchmark.

\subsection{Evaluation, and Statistics}\label{subsec:protocol}
PPO uses fixed hyperparameters across scenarios; training/validation splits use disjoint seeds. Each trained policy is evaluated on \textbf{500} independent day-episodes per scenario. We employ \emph{common random numbers} (paired seeds) across policies and oracles to reduce variance in paired comparisons. For every metric, we report the sample mean $\pm$\emph{(sd)} over the 500 episodes; when stratifying by emergency incidence, results are shown for \emph{All Days}, \emph{Emergency Days}, and \emph{Non-Emergency Days}.

\noindent\textbf{Primary outcome.}
Cumulative reward (\textbf{CR}) $\ \mathbb{E}[\sum_{i\in\mathcal{I}}\Big(u_{k_i}-c_{k_i}\,\omega_i^2\Big)
\;-\;
C_o\cdot \mathrm{OT}]$ (Eq.~\eqref{eq:day_reward})

\noindent\textbf{Secondary outcomes.}
Throughput:\textbf{PT(E)} (completed electives) and \textbf{PT(NE)} (completed non-electives); \textbf{Unserved(NE)}: count of non-elective patients not started within the day; \textbf{OT}:overtime in slots; \textbf{Delay}: waiting time in slots, measured as the mean of absolute elective delay $|\beta_i-\tau_i|$ and non-elective wait $\beta_i-\alpha_i$ over all cases within a day; \textbf{Revenue}: proxy value $\sum_{i\in \mathcal{I} } u_{k_i}$ generated by completed cases.


\subsection{Main Results}\label{subsec:main-results}

\begin{figure}[t]
    \centering
    \begin{subfigure}{0.32\linewidth}
        \centering
        \includegraphics[width=\linewidth]{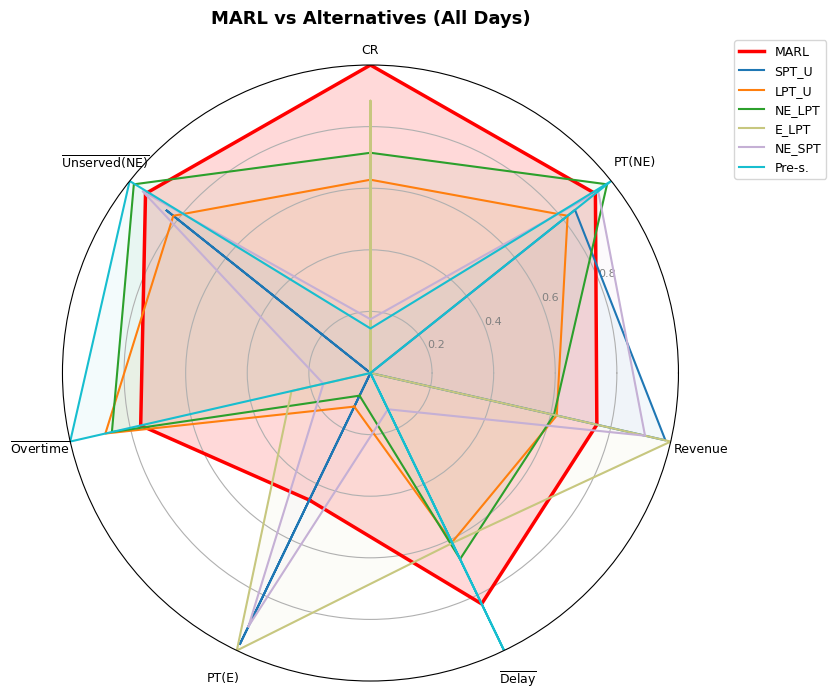}
        \caption{All Days}
        \label{fig:AllDays}
    \end{subfigure}
    \hfill
    \begin{subfigure}{0.32\linewidth}
        \centering
        \includegraphics[width=\linewidth]{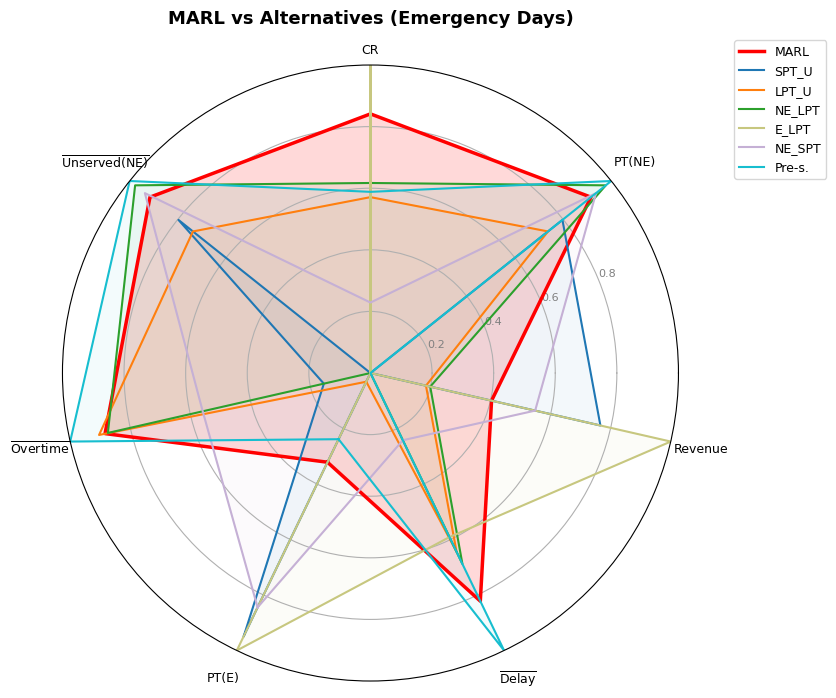}
        \caption{Emergency Days}
        \label{fig:Emergency}
    \end{subfigure}
    \hfill
    \begin{subfigure}{0.32\linewidth}
        \centering
        \includegraphics[width=\linewidth]{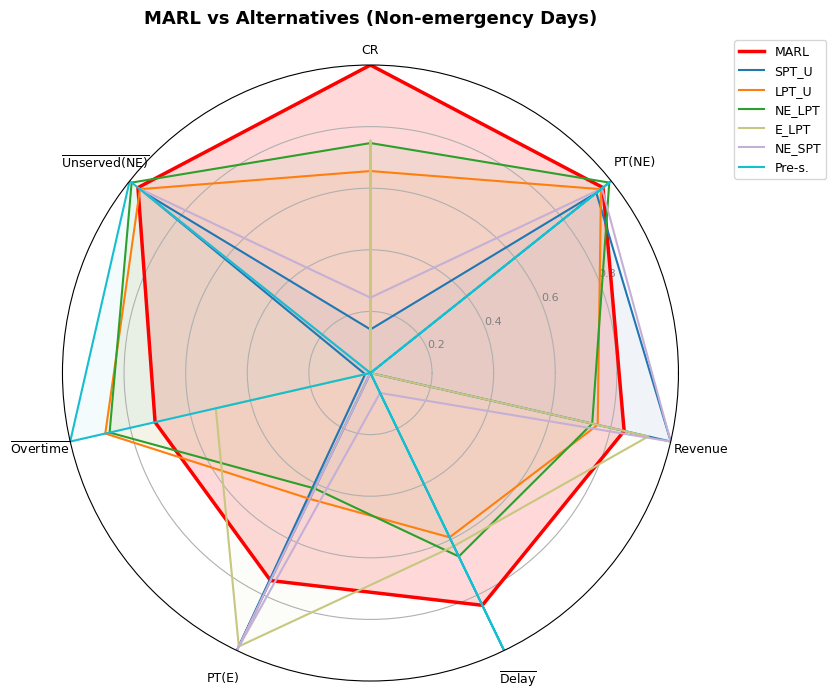}
        \caption{Non-emergency Days}
        \label{fig:NoEmergency}
    \end{subfigure}
    \caption{Radar charts comparing the performance of alternative scheduling methods.}
    \label{fig:radar}
\end{figure}

We compare MARL against all baselines on seven metrics—PT(E), PT(NE), Unserved(NE), OT, Delay, Revenue, and the unified objective CR—using the full test set and the two strata defined by emergency incidence. Raw results appear in Table~\ref{tab:performance comparison}; normalized scores and radar plots are shown in Table~\ref{tab:metrics score comparison} and Fig.~\ref{fig:radar}.

\paragraph{Overall performance.}
Across \emph{All Days}, MARL attains the highest cumulative reward (CR) among all alternative methods, indicating superior overall efficiency. Its variability ($74.28$; Table~\ref{tab:performance comparison}) remains relatively low among alternative methods at a similar performance level (CR above 700), suggesting stable performance. In particular, while \textsc{SPT\_U} ($651.80 \pm 57.81$) and \textsc{SPT\_E} ($672.90 \pm 65.00$) show slightly smaller variance, they do so at substantially lower reward levels. Overall, MARL achieves the best average normalized score ($\mathbf{0.81}$; Table~\ref{tab:metrics score comparison}), reflecting balanced gains rather than single-metric dominance. In the \emph{Emergency Days} stratum, \textsc{E\_LPT} achieves the largest CR but does so by sacrificing non-elective service (low PT(NE), high Unserved) and incurring high overtime (OT). MARL achieves the \emph{best overall balance} (top average score $=\mathbf{0.73}$) with strong CR and materially lower OT/Delay than elective-first rules. In \emph{Non-Emergency Days}, MARL again leads on CR and posts the top average score ($\mathbf{0.87}$), with low Unserved(NE), moderate OT, and low Delay.

\paragraph{Trade-offs.}
Heuristics exhibit clear trade-offs visible in Fig.~\ref{fig:radar}. \textsc{E\_LPT} maximizes PT(E) and Revenue but under-serves non-electives (high Unserved) and often incurs more OT in emergency days; \textsc{SPT\_U} boosts PT(E) yet suffers from OT and Delay, depressing CR. The \textsc{Pre-s} policy controls OT and Delay but leaves reward on the table (lower Revenue/CR). \textsc{NE\_LPT} performs relatively well when no emergency occurs but degrades when an emergency arrives. In contrast, MARL’s polygon is consistently well-rounded: it maintains high PT(NE) and low Unserved while limiting OT and Delay, yielding superior CR.

\paragraph{Policy interpretability.}
Ex-post analytics (Section~\ref{subsec:interpret}) reveal that the learned policy (i) \emph{Batches} compatible types when setup penalties would otherwise accumulate, (ii) \emph{Length-aware}: longer or more complex cases are placed earlier in the day, while shorter ones fill later slots, following an “early-long, late-short” pattern, (iii) \emph{Adaptive prioritization}: the policy raises the priority of urgent and emergency cases, inserting them adaptively, and (iv) \emph{Load balancing} long or complex cases are staggered across operating rooms, preventing simultaneous congestion and promoting balanced utilization throughout the system. These patterns align with our unified reward design and provide face-valid, managerial explanations for the balanced performance observed in Fig.~\ref{fig:radar}.

\paragraph{Regret Analysis.}
To assess how closely the learned policy approaches the theoretical upper bound, we conduct a regret analysis relative to an oracle benchmark with full knowledge of future arrivals and realized durations (Section~\ref{subsec:oracle_regret}) .
The oracle represents an unattainable but informative performance ceiling, and the difference in objective value between the learned policy and the oracle quantifies the regret. Two dominant components account for this gap: (i) the oracle’s knowledge of future arrivals enables timely and precise responses to non-elective cases, substantially reducing waiting-related penalties and service delays, and (ii) with full information about all arrivals and durations, the oracle can globally coordinate the schedule to maximize room utilization and revenue, achieving near-perfect case coverage within capacity limits.

Taken together, the tables and radar plots demonstrate that MARL is \emph{robust and effective} for the multi-objective OR scheduling problem: it dominates or nearly dominates across metrics, delivers the highest unified reward, and preserves service quality for non-electives without incurring excessive overtime—precisely the balanced behavior sought in practice.

\subsection{Computing Facility and Runtime}\label{subsec:compute}
All experiments were conducted on an Apple M2 chip with an 8-core CPU, an 8-core integrated GPU, and 8 GB of unified memory. Training each scenario’s MARL policy required 2.07 hours of wall-clock time over 1,000 epochs, with 16 simulated day-episodes per epoch (i.e., 16 full daily scheduling simulations used for gradient updates). The average simulation throughput was 4,269 environment steps per second. During inference, per-epoch action selection required 0.374 milliseconds on average, and each full-day simulation took 0.192 seconds including the within-epoch sequential assignment.


\begin{table}[ht]
\caption{Performance comparison of the proposed MARL scheduling framework and rule-based scheduling heuristics under all samples and emergency-case samples}
\label{tab:performance comparison}
\resizebox{\textwidth}{!}{
\begin{tabular}{l|*{7}{S[table-format=3.2(4)]}}
\hline
 & \multicolumn{7}{c}{\textbf{All Days}}  \\
\cline{2-8}  
& \text{PT(E)}  & \text{PT(NE)} & \text{Unserved(NE)} & \text{Overtime} & \text{Delay} & \text{Revenue} & \text{CR}\\ \hline
\text{SPT\_U} & 54.07\pm1.33 & 8.80\pm3.12 & 0.74\pm1.38 & 93.67\pm45.51 & 36.20\pm2.07 & 892.72\pm69.78 & 651.80\pm57.81   \\
\text{LPT\_U} & 39.19\pm13.95 &   8.68\pm2.92  &  0.86\pm1.57  &  39.11\pm37.67  &  22.78\pm4.62  &  824.90\pm73.45  &  727.64\pm85.36    \\
NE\_LPT &  38.51\pm14.20  &  9.33\pm3.52  &  0.22\pm0.78  &  40.45\pm37.34  &  21.45\pm4.29  &  823.17\pm72.11  &  738.21\pm80.75    \\
\text{E\_LPT} &  54.49\pm2.18   &  5.48\pm2.67  &  4.07\pm4.23  &  77.42\pm61.18  &  22.73\pm2.67  &  896.03\pm101.95  &  758.64\pm96.19    \\
\text{NE\_SPT} &  52.97\pm3.58  &  9.18\pm3.57  &  0.36\pm0.98  &  83.97\pm41.17  &  33.32\pm3.67  &  879.99\pm63.22  &  672.90\pm65.00    \\
\text{Pre-s.} &  37.09\pm6.99  &  9.39\pm3.59  &  0.15\pm0.67  &  31.87\pm37.66  &  14.21\pm3.35  &  708.67\pm94.95  &  669.28\pm96.72    \\
\text{MARL} &  45.05 \pm 11.19  &  9.14 \pm 3.36  &  0.41 \pm 1.27  &  46.38 \pm 34.47  &  17.91 \pm 3.14  &  849.93 \pm 69.79  &  772.64\pm 74.28  \\ \hline\hline
 &  \multicolumn{7}{c}{\textbf{Emergency Days}} \\
\cline{2-8}  
& \text{PT(E)}  & \text{PT(NE)} & \text{Unserved(NE)} & \text{Overtime} & \text{Delay} & \text{Revenue} & \text{CR}\\ \hline
\text{SPT\_U} &  52.83\pm1.15  &  10.98\pm2.44  &  1.86\pm1.79  &  122.54\pm41.02  &  35.81\pm2.67  &  945.80\pm75.29  &  689.90\pm65.84    \\
\text{LPT\_U} &  24.95\pm9.45  &  10.55\pm2.15  &  2.29\pm1.86  &  67.02\pm46.11  &  21.74\pm6.12  &  848.28\pm85.96  &  774.33\pm85.46    \\
\text{NE\_LPT} &  24.00\pm9.51  &  12.25\pm2.57  &  0.59\pm1.22  &  69.20\pm44.65  &  20.78\pm6.02  &  850.66\pm80.08  &  781.13\pm84.31    \\
\text{E\_LPT} &  54.43\pm2.42  &  5.33\pm2.34  &  7.51\pm3.86  &  134.05\pm60.54  &  22.90\pm2.92  &  985.13\pm104.50  &  837.76\pm106.11    \\
\text{NE\_SPT} &  49.70\pm4.35  &  11.97\pm2.92  &  0.87\pm1.47  &  92.62\pm40.92  &  30.52\pm3.99  &  909.32\pm79.98  &  723.72\pm67.46    \\
\text{Pre-s.} &  31.26\pm6.58  &  12.41\pm2.64  &  0.43\pm1.08  &  59.89\pm47.15  &  13.98\pm3.99  &  817.24\pm56.82  &  776.87\pm61.82  \\
\text{MARL} &  33.78 \pm 11.05   &  11.82 \pm 2.55  &  1.02 \pm 1.93  &  68.50 \pm 43.05  &  17.86 \pm 4.79  &  884.95\pm86.47  &  814.30 \pm 84.41  \\ \hline\hline
 &  \multicolumn{7}{c}{\textbf{Non-emergency Days}} \\
\cline{2-8}  
& \text{PT(E)}  & \text{PT(NE)} & \text{Unserved(NE)} & \text{Overtime} & \text{Delay} &\text{Revenue} & \text{CR}\\ \hline
\text{SPT\_U} &  54.73\pm0.87  &  7.64\pm2.81  &  0.14\pm0.45  &  78.25\pm39.94  &  36.40\pm1.62  &  864.39\pm46.12  &  631.46\pm40.28    \\
\text{LPT\_U} &  46.79\pm9.20  &  7.69\pm2.78  &  0.10\pm0.51  &  24.21\pm20.10  &  23.34\pm3.44  &  812.42\pm62.32  &  702.71\pm74.10   \\
\text{NE\_LPT} &  46.25\pm9.42  &  7.76\pm2.91  &  0.02\pm0.16  &  25.11\pm19.95  &  21.81\pm2.91  &  808.50\pm62.71  &  715.30\pm68.55    \\
\text{E\_LPT} &  54.52\pm2.04  &  5.56\pm2.82  &  2.23\pm3.12  &  47.19\pm34.05  &  22.63\pm2.52  &  848.48\pm60.12  &  716.41\pm55.29   \\
\text{NE\_SPT} &  54.72\pm0.87  &  7.70\pm2.90  &  0.09\pm0.31  &  79.36\pm40.55  &  34.82\pm2.34  &  864.33\pm44.84  &  645.77\pm44.00    \\
\text{Pre-s.} &  40.20\pm4.91  &  7.78\pm2.93  &  0.00\pm0.00  &  16.91\pm18.61  &  14.34\pm2.96  &  650.72\pm49.53  &  611.86\pm53.21  \\
\text{MARL} &  51.07\pm5.11  &  7.71\pm2.89  &  0.08 \pm 0.27  &  34.57\pm21.51  &  17.93\pm1.70  &  831.23\pm50.38  &  750.40\pm57.11  \\ \hline\hline
\end{tabular}}
\end{table}

\begin{table}[ht]
\centering

\caption{Performance comparison of the proposed MARL scheduling framework and rule-based scheduling heuristics under all samples and emergency-case samples}
\label{tab:metrics score comparison}
\resizebox{\textwidth}{!}{
\begin{tabular}{l|cccccccc}
\hline
 & \multicolumn{8}{c}{\textbf{All Days}}  \\
\cline{2-9}  
& PT(E)  & PT(NE) & Unserved(NE) & Overtime & Delay & Revenue & CR & Average Score\\ \hline
SPT\_U & 0.98 & 0.85 & 0.85 & 0.00 & 0.00 & 0.98 & 0.00 & 0.52   \\
LPT\_U & 0.12 & 0.82 & 0.82 & 0.88 & 0.61 & 0.62 & 0.63 & 0.64   \\
NE\_LPT & 0.08 & 0.98 & 0.98 & 0.86 & 0.67 & 0.61 & 0.72 & 0.70   \\
E\_LPT & 1.00 & 0.00 & 0.00 & 0.26 & 0.61 & 1.00 & 0.88 & 0.54  \\
NE\_SPT & 0.91 & 0.95 & 0.95 & 0.16 & 0.13 & 0.91 & 0.17 & 0.60   \\
Pre-scheduled based & 0.00 & 1.00 & 1.00 & 1.00 & 1.00 & 0.00 & 0.14 & 0.59   \\
MARL & 0.46 & 0.93 & 0.93 & 0.77 & 0.83 & 0.75 & 1.00 & 0.81 \\ \hline\hline
 &  \multicolumn{8}{c}{\textbf{Emergency Days}} \\
\cline{2-9}  
& PT(E)  & PT(NE) & Unserved(NE) & Overtime & Delay & Revenue & CR & Average Score\\ \hline
SPT\_U & 0.95 & 0.80 & 0.80 & 0.16 & 0.00 & 0.77 & 0.00 & 0.49   \\
LPT\_U & 0.03 & 0.74 & 0.74 & 0.90 & 0.64 & 0.18 & 0.57 & 0.54   \\
NE\_LPT & 0.00 & 0.98 & 0.98 & 0.87 & 0.69 & 0.20 & 0.62 & 0.62   \\
E\_LPT & 1.00 & 0.00 & 0.00 & 0.00 & 0.59 & 1.00 & 1.00 & 0.51  \\
NE\_SPT & 0.84 & 0.94 & 0.94 & 0.56 & 0.24 & 0.55 & 0.23 & 0.61  \\
Pre-scheduled based & 0.24 & 1.00 & 1.00 & 1.00 & 1.00 & 0.00 & 0.59 & 0.69   \\
MARL & 0.32 & 0.92 & 0.92 & 0.88 & 0.82 & 0.40 & 0.84 & 0.73 \\ \hline\hline
 &  \multicolumn{8}{c}{\textbf{Non-emergency Days}} \\
\cline{2-9}  
& PT(E)  & PT(NE) & Unserved(NE) & Overtime & Delay & Revenue & CR & Average Score\\ \hline
SPT\_U & 1.00 & 0.94 & 0.94 & 0.02 & 0.00 & 1.00 & 0.14 & 0.58   \\
LPT\_U & 0.45 & 0.96 & 0.96 & 0.88 & 0.59 & 0.76 & 0.66 & 0.75   \\
NE\_LPT & 0.42 & 0.99 & 0.99 & 0.87 & 0.66 & 0.74 & 0.75 & 0.77 \\
E\_LPT & 0.99 & 0.00 & 0.00 & 0.52 & 0.62 & 0.93 & 0.75 & 0.54  \\
NE\_SPT & 0.99 & 0.96 & 0.96 & 0.00 & 0.07 & 0.99 & 0.24 & 0.61  \\
Pre-scheduled based & 0.00 & 1.00 & 1.00 & 1.00 & 1.00 & 0.00 & 0.00 & 0.57   \\
MARL & 0.75 & 0.97 & 0.97 & 0.72 & 0.84 & 0.84 & 1.00 & 0.87 \\ \hline\hline
\end{tabular}}
\end{table}

\subsection{Policy Interpretability and Behavioral Insights}\label{subsec:interpret}

\begin{figure}[t]
    \centering
    \subfloat[Non-emergency day]{       \includegraphics[width=0.96\textwidth]{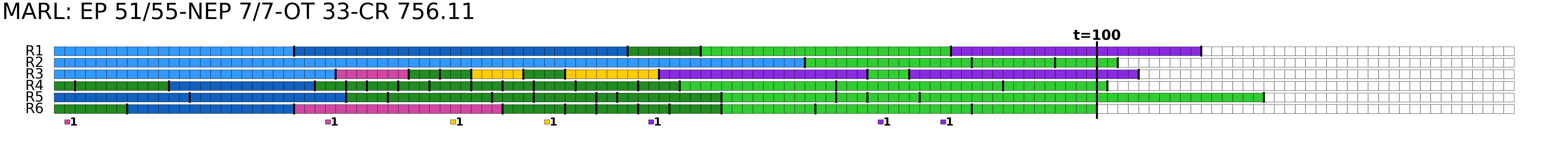}
    }\\[-2pt]
    \subfloat[Late-emergency day ($t=66$)]{       \includegraphics[width=0.96\textwidth]{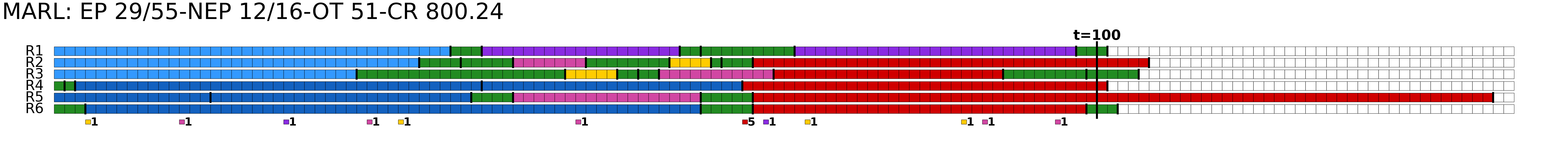}
    }\\[-2pt]
    \subfloat[Early-emergency day ($t=1$)]{
     \includegraphics[width=0.96\textwidth]{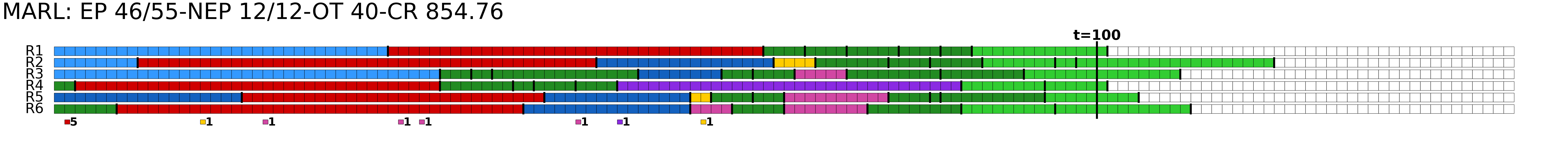}
    }\\[4pt]
    \includegraphics[width=0.96\textwidth]{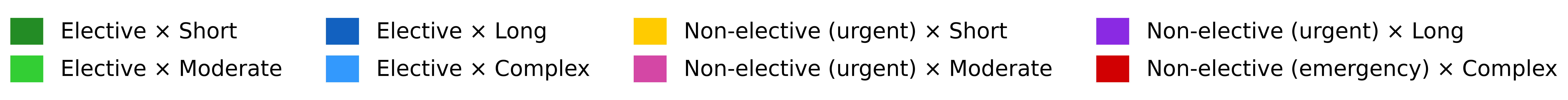}
    \caption{Representative Gantt charts illustrating learned scheduling behavior under different operating conditions:
(a) a regular non-emergency day,
(b) a day with an emergency batch arriving late ($t=66$), and
(c) a day with an early emergency batch ($t=1$) occurring while surgeries are in progress, constraining the MARL policy and leading to longer delays.}
    \label{fig:gantt_examples}
\end{figure}

Beyond aggregate metrics, we visualize representative daily schedules to qualitatively examine how the learned policy behaves under different operational conditions. Figure Figure~\ref{fig:gantt_examples} presents three illustrative Gantt charts corresponding to (a) a regular non-emergency day, (b) a day with an emergency batch arriving late in the schedule ($t=66$), and (c) a day with an early emergency arrival ($t=1$).

These visualizations highlight how the MARL policy adapts its sequencing and resource allocation in response to real-time uncertainty.

\textbf{\emph{Batching compatible types.}} The policy tends to group setup-compatible elective cases—typically shorter procedures-consecutively within each operating room, minimizing setup transition that would otherwise accumulate between heterogeneous case types. 
This batching behavior indicates that the policy has internalized setup-related penalties and implicitly learned to reduce unnecessary switching.

\textbf{\emph{Length-aware scheduling.}} Longer or more complex surgeries are typically placed earlier in the day, while shorter ones occupy later slots, forming an ``early-long, late-short'' temporal structure. 
This pattern mitigates the risk of overtime by ensuring that long cases finish before the end-of-day horizon, while remaining capacity can flexibly absorb shorter cases.

\textbf{\emph{Adaptive prioritization.}} The policy internalizes the prioritization rule directly from the reward structure, raising the priority of urgent and emergency arrivals and inserting them adaptively into the remaining schedule.
This learned behavior enhances responsiveness and ensures timely treatment of high-value or time-critical patients without causing excessive disruption to elective throughput.

\textbf{\emph{Cross-OR load balancing.}} Long or resource-intensive cases are staggered across operating rooms rather than concentrated in a few, preventing simultaneous congestion and promoting balanced utilization system-wide. 
This spatial diversification reduces bottlenecks and improves overall robustness to stochastic variation in case duration.

Taken together, these patterns demonstrate that the MARL policy has learned an operationally coherent strategy. 
The resulting behaviors—batching similar cases, front-loading longer ones, adaptively prioritizing urgent arrivals, and distributing load across ORs—collectively explain the policy’s superior and well-balanced performance across metrics.

\subsection{Regret Analysis Relative to the Oracle}\label{subsec:oracle_regret}

To better understand the remaining performance gap, we compare the learned MARL policy with an ex post MIP oracle that possesses complete foresight of all future stochastic information—namely, the realized surgery durations and the arrival times and types of non-elective patients. The oracle corresponds to the optimal schedule derived from the mixed-integer programming (MIP) formulation, solved to a 1.15\% optimality gap\footnote{The reported optimality gap reflects the solver’s internal tolerance, meaning the MIP solution is guaranteed to be within 1.15\% of the true optimal value of the formulated problem. It measures solver precision, not the difference between the MIP oracle and MARL.}. It therefore represents an idealized yet unrealizable upper bound: computationally near-optimal, but infeasible for real-time deployment.

\begin{figure}[t]
    \centering
    \includegraphics[width=0.95\textwidth]{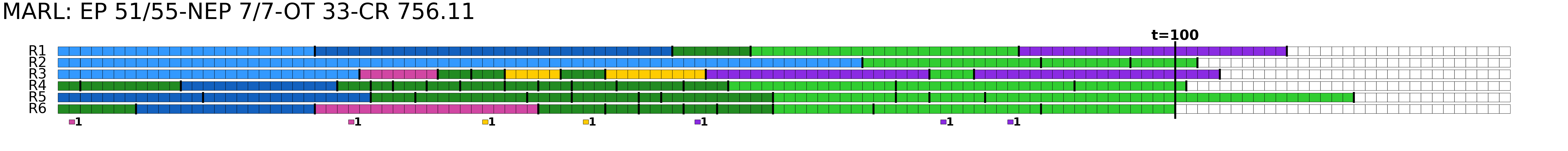}
    \includegraphics[width=0.95\textwidth]{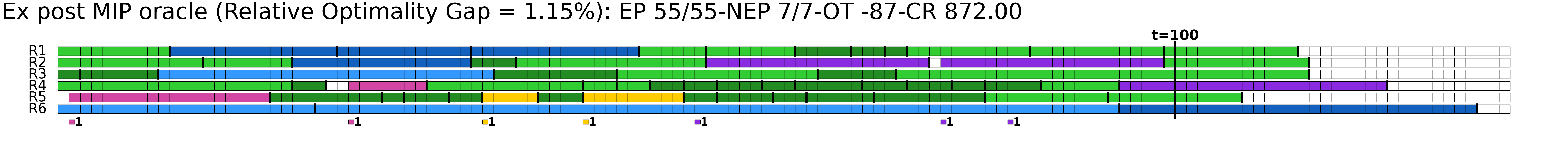}
    \includegraphics[width=0.9\textwidth]{figure/legend.png}

    \caption{Comparison of daily schedules produced by the MARL (top) and the ex post MIP oracle (bottom) under identical arrival sequences.}
    \label{fig:oracle_compare}
\end{figure}

Figure~\ref{fig:oracle_compare} contrasts representative daily schedules generated by the MARL policy (top) and the oracle (bottom) under identical arrival sequences. With perfect foresight, the oracle “looks ahead” and anticipates the moderate-urgency emergency batch, smartly reserving idle capacity in certain operating rooms while maintaining tight coordination across others. This enables it to accommodate all 55 elective and 7 non-elective patients within the scheduling horizon, achieving full coverage and a cumulative reward (CR) of 872.00. By contrast, the MARL policy—operating without advance knowledge—achieves a slightly lower CR of 756.11, primarily due to fewer elective surgeries completed.

This comparison highlights the strategic foresight advantage of the oracle: it can pre-emptively adjust to future shocks, something no real-time policy can replicate. Nevertheless, the MARL policy’s performance remains close to the oracle benchmark, indicating that CTDE learning captures much of the attainable efficiency despite uncertainty and information constraints.

\section{Discussion}\label{sec:discussion}

This section synthesizes managerial implications, limitations, and avenues for future research arising from our study. We also reflect on the theory and on the external validity of our simulation-based evaluation.

\paragraph{Managerial takeaways.}
First, a \emph{unified reward} (Eq.~\eqref{eq:day_reward})—combining throughput/value $u_k$, timeliness via $c_k \omega_i^2$, and overtime $C_o\cdot\mathrm{OT}$—is a practical way to operationalize the multi-objective nature of intraday OR control. In our experiments (Section~\ref{sec:experiments}), the learned policy consistently trades off elective delay and overtime while preserving urgent access (Fig.~\ref{fig:radar}). Second, \emph{interpretable behaviors} emerge without hand-crafted rules: the policy batches shorter and similar case types to minimize setup transitions, forms an “early-long, late-short’’ temporal structure that mitigates overtime risk, dynamically prioritizes urgent or high-$p_k$ arrivals, and staggers resource-intensive cases across operating rooms to prevent congestion and enhance overall system robustness. These patterns align with expert intuition, making the approach more acceptable to clinical stakeholders. Third, the \emph{sequential within-epoch assignment} offers a scalable alternative to exhaustive joint optimization—simple to integrate into a control-desk workflow and fast at runtime—while remaining compatible with centralized training.

\paragraph{What the theory tells us.}
Our Bellman formulation (Eq.~\eqref{eq:bellman}) confirms the problem is a standard finite-horizon MDP; optimal joint planning is, however, computationally prohibitive at realistic scales. Under weak coupling (Lemma~\ref{lem:weak-coupling}, assumptions A1–A4), the sequential construction is \emph{equivalent} to joint action selection; outside that regime, the gap is driven by \emph{externalities} across rooms. The regret bounds (Eqs.~\eqref{eq:regret_wait}–\eqref{eq:regret_T_bound} and Corollary ~\ref{cor:urgent}) show that loss relative to a centralized optimum concentrates in two channels: excess waiting of high-penalty patients and terminal overtime. Empirically, these are exactly the components in which heuristics underperform MARL.

\paragraph{Scope and limitations.}
\emph{(i) Homogeneous ORs and missing staff couplings.}
We assume any room can process any type and omit explicit surgeon/anaesthesia/nursing rosters or PACU/ICU capacity. This abstraction matches many day-of-control desks but overlooks cross-resource conflicts, break rules, and shift relief. Section~\ref{sec:formulation} assumes homogeneous operating rooms with identical capabilities. Extending the framework to heterogeneous or coupled resources is a natural direction for future work.
\emph{(ii) FIFO within type.}
Within-type FIFO simplifies queue discipline; urgency is encoded via $\{u_k,c_k\}$ rather than hard priorities. Hospitals that enforce strict clinical triage can incorporate service-level constraints or minimum-priority rules into the reward or as hard constraints.

\emph{(iii) Arrival and duration models.}
Non-elective demand is exogenous and modeled by Poisson/Bernoulli processes; durations use Gamma/mixtures fitted to empirical moments. These choices are standard and face-valid but imperfect: true processes can be nonstationary (seasonality, special clinics) and exhibit heavier tails or surgeon effects. Deployment should re-calibrate $\{\lambda_k(t)\}$ and $\mathcal{D}_k$ from local data.

\emph{(iv) Reward calibration.}
Weights $(u_k,c_k,C_o)$ bundle clinical value, timeliness, and labor costs (including staff well-being). They should be co-designed with stakeholders and finance to reflect local priorities and compliance targets; different choices can tilt policies toward access or overtime.

\emph{(v) Oracle comparators.}
We benchmark against ex post oracles solely to quantify regret and headroom; these are \emph{unfair} and not deployable. 

\emph{(vi) Reproducibility and data access.}
Due to NDAs, we release a synthetic generator calibrated to empirical moments rather than raw hospital data. We provide full hyperparameters, seeds, and code for the simulator and training in the e-companion to facilitate replication with local data.

\paragraph{Threats to validity.}
The evaluation covers 500 independent days, but it remains a simulation study. Distribution shift—e.g., calendar effects, surgeon-specific variability, or unplanned staffing outages—can degrade performance. Offline evaluation using retrospective logs and shadow-mode A/B pilots are important intermediate steps before live deployment. Additionally, while PPO with shared actor–critic performed robustly, learning dynamics can be sensitive to network size, entropy regularization, or advantage estimation.

\paragraph{Scalability and compute.}
The joint action space grows exponentially in $J$, but our CTDE + sequential protocol scales linearly in the number of available rooms per epoch. Inference is fast (milliseconds per epoch on commodity hardware; Section~\ref{subsec:compute}); training is offline and can be amortized. Larger suites (e.g., $J>12$) or finer time discretization primarily increase simulation cost rather than action complexity.

\paragraph{Ethical and operational considerations.}
Policies that “optimize” overtime may inadvertently shift burdens across staff. Our framework can incorporate explicit \emph{staff well-being} terms (e.g., soft caps on extended shifts) via additional penalties or constraints. Likewise, equity across patient groups (e.g., elective specialities) can be handled via minimum service levels or fairness-aware weights. Interpretability is key for adoption; post-hoc policy explanations should accompany any rollout.

\paragraph{Future work.}
\emph{(A) Coupled resources and downstream units.} Extend the state to include surgeon/anaesthesia rosters, nursing, and PACU/ICU/bed capacity; consider multi-resource setups and handoffs.  
\emph{(B) Constrained and risk-sensitive RL.} Enforce service-level constraints (e.g., maximum non-elective wait quantiles), Conditional Value-at-Risk (CVaR) based overtime control, or chance constraints on late finishes.  
\emph{(C) Stronger theory.} Move beyond weak-coupling equivalence to tighter bounds using submodularity or matroid-intersection structure; analyze convergence of CTDE under sequential action composition.  
\emph{(D) Generalization and adaptation.} Meta-learning across hospitals, context-conditioned policies, distributionally robust training to handle nonstationarity, and stochastic disruptions such as last-minute surgery cancellations.  
\emph{(E) Policy distillation.} Extract compact, human-readable rules (e.g., decision trees) from the learned policy to facilitate audit and governance.  
\emph{(F) Human-in-the-loop integration.} Embed the policy in an OR dashboard that proposes actions, supports “what-if’’ simulation, and allows supervisor overrides with learning from corrections.

\smallskip
In sum, MARL with a unified reward, centralized training, and sequential execution offers a pragmatic, data-driven complement to optimization-based control. It delivers balanced performance across competing objectives, scales to realistic suites, and yields interpretable behavior—all prerequisites for impact in intraday OR scheduling. The limitations above are tractable engineering and modeling extensions rather than fundamental barriers, and we view this work as a step toward robust, clinician-aligned decision support at the OR control desk.

\section{Conclusion}\label{sec:conclusion}

Intraday operating room scheduling is a high-stakes, multi-objective problem that must balance timely access for urgent patients, elective timeliness, and overtime exposure under substantial uncertainty. We formulated this problem as a Markov game with a \emph{unified reward} (Eq.~\eqref{eq:day_reward}) that internalizes throughput/value, delay, and overtime into a single, optimizable signal. Building on this foundation, we proposed a CTDE multi-agent reinforcement learning approach that uses a shared actor–critic and a \emph{within-epoch sequential assignment} to produce conflict-free joint actions without enumerating the exponential joint-action space. The pre-day MIP plan provides elective reference times that anchor delay penalties while remaining nonbinding intraday.

On the theory side, we established a dynamic-programming basis (Eq.~\eqref{eq:bellman}) and identified conditions under which the sequential construction is equivalent to joint action selection (Lemma~\ref{lem:weak-coupling}). Beyond this weak-coupling regime, we derived a bound that decomposes regret into excess waiting and overtime (Eqs.~\eqref{eq:regret_wait}–\eqref{eq:regret_T_bound}) and provided a clean gap result in a stylized urgent-arrival case (Corollary ~\ref{cor:urgent}). These results clarify where performance losses can arise and motivate diagnostics and guardrails.

Empirically, across 500 independent day-episodes, the learned policy consistently delivers the highest unified reward (CR), with robust improvements in urgent access, elective delay, and overtime relative to a suite of transparent heuristics. Oracle analyses---while \emph{unfair} due to ex post information---show modest, interpretable gaps concentrated in the theoretically identified channels. Policy analytics reveal face-valid behaviors (type batching, length-aware scheduling, adaptive prioritization and cross-OR load balancing, supporting managerial acceptability.

The approach is practical: it scales linearly with the number of available rooms per epoch, runs quickly at inference, and offers interpretable summaries of learned behavior. Limitations---homogeneous ORs, omitted staff couplings, and simplified arrival/duration models---are addressable through extensions discussed in Section~\ref{sec:discussion}, including coupled-resource modeling, constrained or risk-sensitive RL, and human-in-the-loop deployment.

Overall, MARL with sequential assignment is a viable, data-driven complement to optimization for real-time OR scheduling. By unifying objectives and learning coordinated policies directly from realistic simulations, it achieves balanced performance difficult to obtain with fixed heuristics, while preserving the transparency and scalability needed for clinical operations. We view this work as a step toward reliable decision support at the OR control desk and a foundation for future advances in coupled-resource scheduling, fairness, and safe learning from real hospital data.

\bibliography{References_new}

\end{document}